\documentclass[sigconf,10pt, nonacm
]{acmart}
\usepackage{booktabs}
\usepackage{array}
\usepackage{graphicx}
\graphicspath{{figures/}}
\usepackage{xcolor}
\usepackage{url}
\usepackage{enumitem}
\usepackage{tikz}
\usetikzlibrary{arrows.meta,positioning}
\usepackage{pifont}
\usepackage{colortbl}
\definecolor{hostfill}{HTML}{E3F0F2}
\definecolor{hostteal}{HTML}{2B7A8C}
\definecolor{badred}{HTML}{D1495B}
\newcommand{\nd}{\textcolor{black!35}{---}}
\definecolor{amber}{HTML}{C7801F}\definecolor{amberdark}{HTML}{9A5F0E}\definecolor{amberfill}{HTML}{FBF4E8}
\usepackage[most]{tcolorbox}
\usepackage{fontawesome5}
\definecolor{gridgiven}{HTML}{CFCFCF}\definecolor{okfill}{HTML}{CFE6EA}
\newtcolorbox{takeaway}[3]{enhanced, colback=#2, colframe=#1, boxrule=0pt, leftrule=2.5pt, arc=1.5pt, left=6pt,
  right=6pt, top=4pt, bottom=4pt, fonttitle=\bfseries, coltitle=black!88, title={#3}, attach title to upper,
  after title={\par\smallskip}}
\AtBeginDocument{\hypersetup{colorlinks=true, linkcolor=hostteal, citecolor=hostteal, urlcolor=hostteal}}
\renewcommand\footnotetextcopyrightpermission[1]{}
\acmConference[MobiSys '27]{The 25th Annual International Conference on Mobile Systems, Applications and Services}{June 21--25, 2027}{Ho Chi Minh City, Vietnam}
\acmYear{2027}
\copyrightyear{2027}
\setcopyright{none}

\newcommand{\methodname}{\textcolor{hostteal}{\texttt{FirstLight}}}

\makeatletter
\g@addto@macro\@mkauthors{\global\setbox\mktitle@bx=\vbox{\unvbox\mktitle@bx\vskip6pt}}
\makeatother
\begin{document}
\title{Intelligence per Iteration: Auditing the Cost of Looped Models on Edge Devices}
\author{Steven Kolawole}
\authornote{Corresponding author: skolawol@cs.cmu.edu.}
\affiliation{\institution{Carnegie Mellon University}\country{}}
\author{Pearse Jim}
\affiliation{\institution{ML Collective}\country{}}
\author{Opegbemi M. Busoye}
\affiliation{\institution{ML Collective}\country{}}
\author{Virginia Smith}
\affiliation{\institution{Carnegie Mellon University}\country{}}

\begin{abstract}
On devices with limited memory, models increasingly gain capability by iterating rather than by growing. Looped models do so most directly, applying one block of weights repeatedly to their hidden state.
However, existing measures cannot tell how such a model performs in practice: parameter count ignores how many iterations it runs, FLOPs ignore the device, and a compiled model that runs is assumed to be correct. In this work, we study seven looped models on 47 devices, from microcontroller NPUs to datacenter GPUs. We find that a model's intelligence per joule on a device rests on the energy, count and correctness of its iterations, and that a few iterations on the device reveal all three. Given the model's intelligence per iteration, which a host measures once, we use the energy of a single iteration to predict the cost of every run and its duration to predict the model's accuracy within any latency budget, whereas FLOP estimates miss the cost by up to $97\times$. Correctness, the third property, cannot be assumed: 94 of our 280 builds run without error yet compute wrong answers. To check it, we compare the hidden state with the host's after two iterations, and catch every held-out broken build without labels. Building on these findings, we propose \methodname{}\footnote{We will release the tooling and benchmark upon publication.}, a simple procedure for deploying one looped model across many devices: it measures the model once on a host, then prices, checks and repairs its build on each device in a few iterations.

\end{abstract}
\maketitle
\section{Introduction}
\label{sec:intro}

\begin{figure}[t]
\centering
\includegraphics{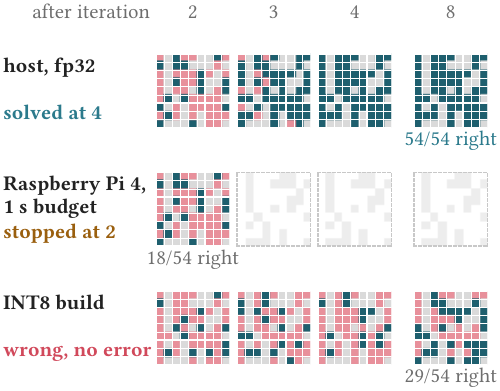}
\caption{\textbf{The same weights fill 54, 18 or 29 of a Sudoku's 54 blank cells correctly, depending on where they run.} Sudoku-attn on one test puzzle; teal cells are right, red ones wrong. The Pi row shows the host's answers after the two iterations that fit in 1\,s; the INT8 build runs under TensorRT on an L40S.}
\label{fig:overview}
\end{figure}

\begin{figure*}[t]
\centering
\includegraphics{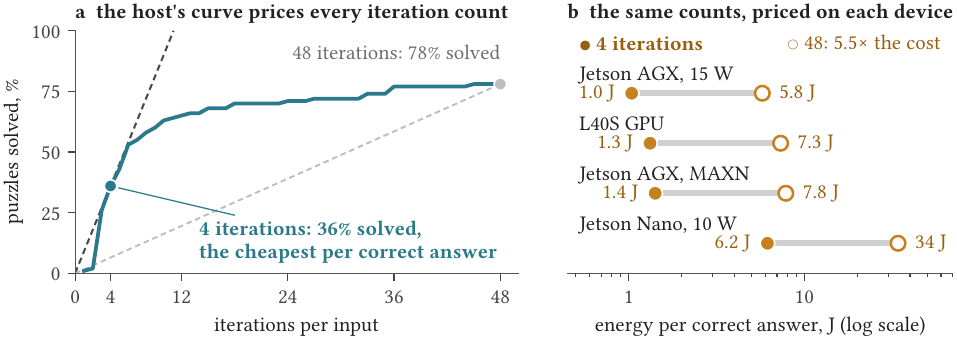}
\caption{\textbf{The host's accuracy curve sets a looped model's cheapest iteration count, 4 for Sudoku-MLP, and it is the same on every device.} (a)~Sudoku-MLP on the host; a ray's slope is the intelligence per iteration. (b)~Energy per correct answer at 4 and at 48 iterations, priced from the energy of a single iteration on each device.}
\label{fig:cost}
\end{figure*}

On devices with limited memory, iteration is increasingly how models gain accuracy: diffusion language models refine their output over many steps~\citep{nie2025llada,dllmnpu2026}, phones and edge GPUs scale test-time compute~\citep{hao2025ttsnpu,fasttts2025}, and on-device agents call a small model in a loop~\citep{tian2026agentsedge}. In each case, the number of iterations sets both how accurate an answer is and what it costs in time and energy, which an edge device can spend only until its battery drains or it throttles. Among these, looped models build the iteration into the network itself: one block of weights runs many times on the model's own hidden state~\citep{geiping2025huginn,bae2025mor,huang2026eqr}. Because the same weights run at every iteration, a looped model grows more accurate without growing larger. Instead, it spends compute, which hardware has scaled far faster than memory~\citep{gholami2024memorywall}. With respect to performance, a looped language model matches transformers up to $3\times$ its size~\citep{zhu2025ouro}, and the Tiny Recursive Model (TRM) solves Sudoku and maze puzzles that far larger language models fail, with 7M parameters~\citep{jolicoeur2025trm,wang2025hrm}. The practical implication is that models this small fit on local hardware, where they answer without a network round trip and keep data on the device, and open-weight looped models now ship at sizes meant for such devices~\citep{nanbeige2026loop,sensenova2026loop,tml2026inkling2mini}. Deploying one, however, raises three questions that apply to every iterating model: what an iteration costs on the device, how many iterations each input runs, and whether the device computes them correctly (Figure~\ref{fig:overview}).

Parameter count and FLOPs, the common measures of efficiency, cannot answer the first two questions, because they describe a network that runs once. Parameter count is the same at 1 iteration or 64, although 64 iterations cost $64\times$ the energy of one. FLOPs are the same on every device, although an iteration of our Sudoku model takes $4$\,ms on a Jetson AGX and $378$\,ms on a Raspberry Pi~4. Modern benchmarks like intelligence per watt avoid these proxies by measuring accuracy per unit of power on the device~\citep{saadfalcon2025ipw}, but they give one point per model and device, while a looped model's accuracy and energy change with every iteration. More broadly, attributing energy to such multistep computation remains an open problem~\citep{tschand2026powerwall}. A developer therefore cannot tell what a looped model will cost on a device, or whether it will answer in time, without running it there at every iteration count.

Running the model on the device does not settle the third question either. A deployment toolchain compiles the model into a \emph{build} for one device at one precision and reports only whether compilation succeeds and the build runs, so a build that runs is taken to be correct. For example, a quantized build of the maze-solving TRM solves $84\%$ of mazes when simulated on a host computer, yet on a Snapdragon NPU it ran without an error and solved none, although most cells of its paths were right. The fault repeats across devices: the standard w8a8 build of the same model is broken on 10 of 13 Snapdragon devices. Every toolchain we checked produced such builds, and a host simulation finds nothing wrong with most of them (\S\ref{sec:faults}). Worse, a broken build keeps drawing power while it answers nothing correctly, so all the energy it spends is wasted. Thus, with no labels, a deployed device cannot notice.

In this work, we study the three questions on seven looped models from five families, among them a looped language model, on 47 devices of 13 types, from microcontroller NPUs to datacenter GPUs (\S\ref{sec:method}). We find that a few iterations on a device answer all three, because every iteration repeats the same computation on a hidden state of the same size. Each iteration therefore costs the same energy, so we can price a run of any length from a single iteration. Likewise, a correct build computes the host's function up to small numerical error, so its accuracy follows the host's at every iteration. A single host run thus gives every device's accuracy curve and the iteration at which each input halts. A broken build instead computes a different function, so its hidden state departs from the host's within the first iterations.

Energy and accuracy together give a looped model's \emph{intelligence per joule} on a device~\citep{saadfalcon2025ipw}, the correct answers it gives per joule. Its inverse, the energy per correct answer, is the model's \emph{cost} on that device, and it grows with the energy and number of iterations and with every wrong answer. Since energy and accuracy both change with the iteration count, intelligence per joule is a curve over iterations, where intelligence per watt gives a single point. Because every iteration costs the same, this curve is the model's \emph{intelligence per iteration} (its correct answers per iteration) divided by the energy of one iteration on the device (Figure~\ref{fig:cost}). We price every point of this curve from the energy of a single iteration, far more closely than a FLOP count can (\S\ref{sec:cost}). Since that energy scales the cost of every iteration count alike, the count at which an answer is cheapest is the same on every device. We therefore set it once, on the host. From the time of an iteration, we then read each device's accuracy within an application's \emph{latency budget}, the time it allows for one answer. For example, within $100$\,ms, the same Sudoku weights solve $71\%$ of puzzles on a Jetson AGX and at most $2\%$ on most phone CPUs (Figure~\ref{fig:ladder}, \S\ref{sec:predict}). Halting, finally, stops each input once the model's learned signal says its answer is ready~\citep{graves2016act,banino2021pondernet}, so easy inputs cost less than hard ones (\S\ref{sec:halting}).

All of this assumes that the device computes each iteration correctly. Since each iteration carries a fault forward to the next (\S\ref{sec:early}), the hidden state shows within two iterations whether it does. By comparing it with the host's after the second iteration, we separate every broken build from every correct one without labels (\S\ref{sec:check}). A build that fails can still be used, because running a few more iterations at higher precision after its run restores its accuracy (\S\ref{sec:repair}).

We combine these findings in \methodname{}, a simple procedure for deploying one looped model on many devices. \methodname{} runs the model once in fp32 on a host; each device then runs its build for two iterations to check and time it, and a build that fails is repaired with a few higher-precision iterations (\S\ref{sec:procedure}). A developer thus learns what each answer will cost, how accurate it is within a latency budget, and whether the build is correct, without running every iteration count on every device or labelling any output. Our contributions are:
\begin{enumerate}[leftmargin=1.4em,itemsep=1pt,topsep=2pt]
  \item \textbf{We characterize looped models on edge hardware} by the energy, count and correctness of their iterations, across seven models, 47 devices and 280 builds. Halting cuts the energy per correct answer by up to $2.3\times$ against the best fixed iteration count, chosen with the test labels (\S\ref{sec:method}, \S\ref{sec:characterize}).
  \item \textbf{Next, we show that the energy of a single iteration is enough to price a looped model on a device,} with a median error of at most $8\%$, where FLOP estimates miss by up to $97\times$. The cheapest iteration count, which maximizes intelligence per iteration, is the same on every device, so the host sets it once (\S\ref{sec:cost}, \S\ref{sec:predict}).
  \item \textbf{We identify silent faults in deployed looped models:} 94 of 280 builds from three toolchains run without error but compute wrong answers, and host simulation misses most of the broken phone builds (\S\ref{sec:faults}, \S\ref{sec:early}).
  \item \textbf{We then show that the hidden state after two iterations is enough to reveal these faults without labels,} with no miss and no false alarm on 72 held-out phone builds and on a held-out GPU (\S\ref{sec:check}).
  \item \textbf{Finally, we propose \methodname{},} which checks a build on each device in two iterations per input, under $1$\,s even on a Raspberry Pi~4, prices it from the energy of a single iteration, and restores a broken build from $0\%$ to the host's $88\%$ (\S\ref{sec:repair}, \S\ref{sec:procedure}).
\end{enumerate}

\section{Background and related work}
\label{sec:background}

\begin{table*}[t]
\centering\setlength{\tabcolsep}{5pt}
\caption{\textbf{Prior work prices one pass of a model and flags drift from a reference; we price every iteration count and check, without labels, whether a build keeps its accuracy.} \nd: not addressed.}
\label{tab:related}
\begin{tabular}{@{}>{\raggedright\arraybackslash}p{0.3\textwidth}>{\centering\arraybackslash}p{0.1\textwidth}>{\raggedright\arraybackslash}p{0.15\textwidth}>{\raggedright\arraybackslash}p{0.14\textwidth}>{\raggedright\arraybackslash}p{0.21\textwidth}@{}}
\toprule
Work & Device energy measured? & Which iteration counts are priced? & Halting compared with & What the build check detects \\
\midrule
\multicolumn{5}{@{}l}{\emph{Cost of inference}} \\
\quad FLOP counts~\citep{yang2017energyaware,fernandez2023frameworktax} & \nd & one pass & \nd & \nd \\
\quad nn-Meter~\citep{zhang2021nnmeter} & \nd & one pass & \nd & \nd \\
\quad InstMeter~\citep{liu2026instmeter} & \checkmark & one pass & \nd & \nd \\
\quad Intelligence per watt~\citep{saadfalcon2025ipw} & \checkmark & one point & \nd & \nd \\
\multicolumn{5}{@{}l}{\emph{Iteration count}} \\
\quad Early exit~\citep{teerapittayanon2017branchynet,laskaridis2020spinn,huang2023fleet} & \checkmark & \nd & the final exit & \nd \\
\multicolumn{5}{@{}l}{\emph{Correctness}} \\
\quad Converter, compiler studies~\citep{jajal2024converters,louloudakis2023deltann} & \nd & \nd & \nd & faults in single-pass models \\
\quad ML-EXray, DiFR, Ekka~\citep{qiu2022mlexray,karvonen2025difr,gu2026ekka} & \nd & \nd & \nd & drift beyond numerical noise, or an accuracy drop on labelled data \\
\midrule
\rowcolor{hostfill}\textbf{\textcolor{hostteal}{This paper}} & \textcolor{hostteal}{\checkmark} & \textcolor{hostteal}{every count, from one iteration (\S\ref{sec:cost})} & \textcolor{hostteal}{the best fixed count (\S\ref{sec:halting})} & \textcolor{hostteal}{lost accuracy, from two iterations, no labels (\S\ref{sec:check})} \\
\bottomrule
\end{tabular}
\end{table*}
\label{sec:related}

In this section, we describe looped models, define how we measure the energy, count and correctness of their iterations, and review prior work on each (Table~\ref{tab:related}).

\textbf{Looped models.}
A looped model applies one block of weights many times to its own hidden state and reads its answer from that state after the last iteration. The state has the same size at every iteration, so every iteration runs the same computation on the same amount of data. For the models we study, accuracy rises with the number of iterations and then levels off (Figure~\ref{fig:cost}a).

The idea predates its recent use for reasoning. DeepThinking networks repeat a convolutional block and solve larger mazes than they were trained on by running more iterations~\citep{schwarzschild2021deepthinking,bansal2022deepthinking}, and multiscale deep equilibrium models (MDEQ) repeat their block until its output no longer changes~\citep{bai2020mdeq}. TRM~\citep{jolicoeur2025trm} and the Hierarchical Reasoning Model (HRM)~\citep{wang2025hrm} carry the design to Sudoku and maze puzzles with a few million parameters. Their repeated block, one high-level cycle, updates one part of the hidden state several times and then a second part once. Looped language models, Ouro among them, repeat their layers for each token to spend more computation on it~\citep{geiping2025huginn,bae2025mor,zhu2025ouro}, and MELT gives Ouro a key-value cache whose size no longer grows with the number of loops~\citep{vendrell2026melt}. These works train looped models and report their accuracy; the nearest to deployment, LoopDuet, serves looped language models on server GPUs for throughput~\citep{du2026loopduet}. None of them measures, on edge devices, what a looped model's iterations cost, how many each input needs, or whether the device computes them correctly.

\textbf{The cost of inference on devices.}
Knowing what inference costs on a device tells a developer which model it can run, how fast it answers and how long it runs before its battery drains or it throttles. Measurement studies characterize models on phones and their NPUs~\citep{almeida2021smart,reddi2022mlperfmobile,li2026npunotfaster}, runtimes such as llm.npu speed them up there~\citep{xu2025llmnpu}, and MELTing point~\citep{laskaridis2024melt}, ML.ENERGY~\citep{chung2025mlenergy} and MLPerf Power~\citep{tschand2025mlperfpower} measure the energy of inference across models and devices. Because measuring every model on every device is expensive, predictors estimate this cost instead. FLOP counts predict neither latency nor energy reliably~\citep{yang2017energyaware,fernandez2023frameworktax,fernandez2025energy,gholami2024memorywall}, and a server's rated power misstates its measured draw~\citep{newkirk2025h100power}. nn-Meter therefore predicts the latency of a forward pass from its kernels~\citep{zhang2021nnmeter}, and InstMeter its energy from its compiled instructions~\citep{liu2026instmeter}. Two lines of work relate this cost to what a model produces. Token-energy models follow a language model's cost as its cache grows with every token~\citep{vellaisamy2026tokenenergy}, and intelligence per watt divides accuracy by power to compare local models and devices~\citep{saadfalcon2025ipw}. These works price one forward pass of a model on a device, compare models at one setting, or follow a cost that grows with every token.

A looped model fits none of these: every iteration costs the same, while its accuracy changes with the number of iterations it runs. We therefore price an answer by its \emph{energy per correct answer},
\begin{equation}
\text{energy per correct answer} = \frac{\text{energy spent on the inputs}}{\text{inputs answered correctly}},
\label{eq:jpc}
\end{equation}
which charges a model both for the iterations it runs and for the inputs it gets wrong. It is the model's cost on the device, as defined in Section~\ref{sec:intro}, and its inverse is the model's intelligence per joule. To separate what the model contributes to this cost from what the device does, we define the model's \emph{intelligence per iteration} as $\mathrm{IPI}(n) = \mathrm{acc}(n)/n$, for accuracy $\mathrm{acc}(n)$ after a mean of $n$ iterations per input. It depends only on the model and its inputs. Since every iteration costs the same, dividing it by the energy of one iteration on a device gives that device's intelligence per joule. In Section~\ref{sec:cost}, we show that the energy of a single iteration is enough to predict a device's intelligence per joule at every iteration count, where FLOP counts miss by up to $97\times$. In Section~\ref{sec:predict}, we show that the time of that iteration gives the accuracy within any latency budget.

\textbf{Halting and early exit.}
A looped model need not run the same number of iterations for every input. A model trained to \emph{halt} predicts, after each iteration, whether its answer is ready, and stops when it is~\citep{graves2016act,banino2021pondernet}. TRM, HRM and Ouro learn such a signal during training, and Energy-Based Reasoners stop once the gradient of a learned energy falls below a threshold~\citep{chen2026ebr}. Early exit makes the same decision at chosen layers of an ordinary network: classifiers attached to intermediate layers stop an input at the first one that is confident~\citep{teerapittayanon2017branchynet,kaya2019sdn,laskaridis2021earlyexit}. Later work splits this computation between a device and a server or between IoT devices and the cloud~\citep{laskaridis2020spinn,huang2023fleet}, and exits per token in language models~\citep{schuster2022calm}. Test-time scaling makes the same trade on phones and edge GPUs, spending more computation on harder inputs by sampling and searching over tokens~\citep{fasttts2025,hao2025ttsnpu}. On server GPUs, continuous depth batching raises the throughput of looped language models that halt per token, and notes that such speedups are often estimated from FLOPs alone~\citep{schwethelm2026cdb}. These studies report their savings per inference, in FLOPs or in throughput, against the network's final exit, a fixed budget or another split of the work.

For a looped model, the natural baseline is instead a fixed number of iterations of the same weights. We compare halting with the cheapest fixed count of the same weights that reaches the same accuracy, chosen with the test labels as an oracle would. In Section~\ref{sec:halting}, we show that halting cuts the energy per correct answer by up to $2.3\times$ against this baseline, and trace each saving to the inputs that halt early.

\textbf{Deployment toolchains and their faults.}
A model reaches a device through the vendor's toolchain, which converts its graph, fuses operators, chooses kernels and usually quantizes it. We study the toolchains that vendors ship for each kind of device: Qualcomm QNN for Snapdragon NPUs~\citep{qualcomm2026qairt}, TensorRT for NVIDIA GPUs~\citep{nvidia2026tensorrt}, ONNX Runtime for CPUs~\citep{onnxruntime2026}, and TFLite and TFLite Micro for phones and microcontrollers~\citep{david2021tflm}. Each of these steps can change what the model computes, even for a model that runs once: converters and compilers introduce faults~\citep{jajal2024converters,shen2021compilerbugs,guan2023optbugs,liu2023nnsmith,chen2021deployfaults}, and the same model computes different outputs on different runtimes and hardware~\citep{louloudakis2023deltann,schlogl2023deviations,shin2026int8portable}. A toolchain does not report what a build computes, so checking it needs a reference.

ML-EXray compares the per-layer outputs of an edge deployment with a reference~\citep{qiu2022mlexray}, and DiFR and Ekka verify a cloud model provider and serving engines~\citep{karvonen2025difr,gu2026ekka}; each flags departures from a trusted implementation beyond numerical noise, and DiFR detects ones as small as 4-bit quantization. A quantized build, however, departs from its reference by design, so such a check cannot tell a build that keeps its accuracy from one that has lost it. We therefore judge a build by the accuracy it keeps against the same model run in fp32 on the same inputs. A build is \emph{correct} when it keeps ${\geq}95\%$ of that accuracy, a margin for small numerical differences, \emph{broken} when it keeps ${<}50\%$, and \emph{degraded} in between.

In Section~\ref{sec:check}, we show that comparing the hidden state after two iterations with an fp32 run separates every broken build from every correct one, where agreement of the outputs, a test for constant output and a host simulation of the quantized graph fall short. Closest to our check, Jim et al.\ use the cosine between the final hidden states of a compressed model and an fp32 one as a label-free signal of compression damage, in simulation~\citep{jim2026survives}; we read it after two iterations of a deployed build on the device, and catch faults that host simulation misses (\S\ref{sec:faults}). Kolawole et al.\ show which compressed looped models keep their accuracy as they iterate, measured on a host~\citep{kolawole2026tilted}; we measure deployed builds on the devices that run them. Quantizers for looped and diffusion models fix the precision or scale of each loop or step offline, on a host~\citep{zhao2026loopquant,fang2026loopq,he2023ptqd,lewandowski2026tmpdiff}, whereas we repair a deployed build by running a few more iterations at higher precision after its run (\S\ref{sec:repair}).

Finally, we study the three properties together, on the devices that run the model. Benchmarks such as intelligence per watt join energy and accuracy~\citep{saadfalcon2025ipw}, but at one point per model and device, and without checking the device's computation against a reference. The other work above prices a model, chooses its iteration count or checks its build, each in isolation and mostly for models that run once. Our work differs by showing that for a looped model, all three properties can be read from a few iterations on the device.

\begin{table*}[t]
\centering
\caption{\textbf{The same weights run on 47 devices of 13 types, where one Sudoku-MLP iteration takes 0.5 to 378\,ms.} Time of one Sudoku-MLP iteration on each type's fastest (dark bar, log scale) and slowest (light) device: kernel time on Qualcomm devices, timed calls on boards and servers, simulation on microcontroller NPUs. \emph{Accuracy}: devices with builds run for accuracy; \emph{energy}: power sensors read. Precisions: TensorRT fp32, fp16, INT8; QNN fp16, w8a8, w8a16, w4a8, w4a16; TFLite fp32; ONNX Runtime fp32, INT8; Ethos-U INT8, w8a16, w4a8. $^{\ast}$\,Not timed for Sudoku-MLP.}
\label{tab:devices}
\setlength{\tabcolsep}{5pt}
\begin{tabular}{@{}>{\raggedright\arraybackslash}p{3.6cm}>{\raggedright\arraybackslash}p{6.5cm}r@{\hspace{4pt}}lcc@{}}
\toprule
Type & Devices & \multicolumn{2}{l}{ms per iteration (log scale)} & Accuracy & Energy \\
\midrule
\rowcolor{black!6}\multicolumn{6}{@{}l}{\textbf{TensorRT}} \\
\quad Datacenter GPUs (2) & L40S, A100 & 0.5 & \textcolor{amber}{\rule{0.10cm}{1.3ex}} & 2 & $\bullet$ \\
\quad Jetson AGX Xavier & AGX Xavier 16 GB & 4.1 & \textcolor{amber}{\rule{0.46cm}{1.3ex}} & 1 & $\bullet$ \\
\quad Jetson Nano (3) & Nano 4 GB ($\times$3) & 63.7 & \textcolor{amber}{\rule{0.94cm}{1.3ex}} & 3 & $\bullet$ \\
\rowcolor{black!6}\multicolumn{6}{@{}l}{\textbf{QNN}} \\
\quad Phones (7) & Galaxy S21$^{\ast}$, Galaxy S22 5G, Galaxy S23, Galaxy S24, Galaxy S25, Galaxy S26, Galaxy A73 5G$^{\ast}$ & 3.1--12.6 & \textcolor{amber}{\rule{0.41cm}{1.3ex}}\textcolor{amber!35}{\rule{0.25cm}{1.3ex}} & 6 & \nd \\
\quad Reference phones (3) & 7 Gen 4 QRD$^{\ast}$, 8 Elite QRD, 8 Elite Gen 5 QRD & 3.2--3.7 & \textcolor{amber}{\rule{0.41cm}{1.3ex}}\textcolor{amber!35}{\rule{0.03cm}{1.3ex}} & 2 & \nd \\
\quad Laptops (3) & X Plus CRD, X Elite CRD, X2 Elite CRD & 3.7--6.1 & \textcolor{amber}{\rule{0.44cm}{1.3ex}}\textcolor{amber!35}{\rule{0.09cm}{1.3ex}} & 1 & \nd \\
\quad IoT boards (5) & Arduino VENTUNO Q, IQ-9075, QCS8550, Q-6690$^{\ast}$, RB3 Gen 2$^{\ast}$ & 5.6--6.5 & \textcolor{amber}{\rule{0.51cm}{1.3ex}}\textcolor{amber!35}{\rule{0.02cm}{1.3ex}} & 3 & \nd \\
\quad Automotive boards (3) & SA7255P ADP, SA8295P ADP, SA8775P ADP & 6.7--15.4 & \textcolor{amber}{\rule{0.54cm}{1.3ex}}\textcolor{amber!35}{\rule{0.15cm}{1.3ex}} & 1 & \nd \\
\rowcolor{black!6}\multicolumn{6}{@{}l}{\textbf{TFLite (the Tab S8: QNN)}} \\
\quad Phones (13) & Pixel 3, Pixel 3a, Pixel 4, Pixel 5, Pixel 6, Pixel 7, Pixel 8, Pixel 9, Pixel 10, Galaxy Note 20, Galaxy A14 5G, Galaxy A53 5G, Redmi Note 10 5G & 18.1--211 & \textcolor{amber}{\rule{0.72cm}{1.3ex}}\textcolor{amber!35}{\rule{0.43cm}{1.3ex}} & \nd & \nd \\
\quad Tablets (2) & Galaxy Tab S7, Galaxy Tab S8$^{\ast}$ & 40.8 & \textcolor{amber}{\rule{0.86cm}{1.3ex}} & 1 & \nd \\
\rowcolor{black!6}\multicolumn{6}{@{}l}{\textbf{ONNX Runtime}} \\
\quad Server CPU & A100 host CPU & 21.9 & \textcolor{amber}{\rule{0.75cm}{1.3ex}} & 1 & \nd \\
\quad Single-board computer & Raspberry Pi 4 & 378 & \textcolor{amber}{\rule{1.25cm}{1.3ex}} & 1 & \nd \\
\rowcolor{black!6}\multicolumn{6}{@{}l}{\textbf{Vela and TFLite Micro}} \\
\quad Microcontrollers (3) & Ethos-U55, U65, U85 & 81.4--111 & \textcolor{amber}{\rule{0.98cm}{1.3ex}}\textcolor{amber!35}{\rule{0.05cm}{1.3ex}} & \nd & \nd \\
\bottomrule
\end{tabular}
\end{table*}

\section{Methodology}
\label{sec:method}

\begin{figure}[t]
\centering
\includegraphics{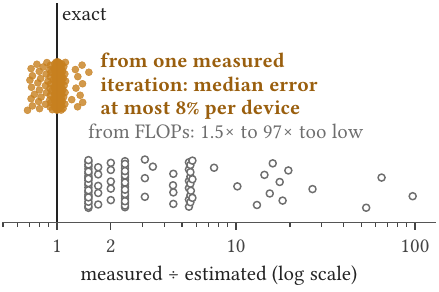}
\caption{\textbf{The energy of a single iteration predicts what a run costs to a median error of at most 8\% on each device; FLOP estimates fall short by up to $97\times$.} One dot per run on the Jetson Nano, Jetson AGX and L40S, each predicted from the energy of an iteration in a separate run with the same model, precision and power mode.}
\label{fig:energy}
\end{figure}

To measure the three properties of iteration, we run seven looped models on 47 devices, with the same weights on every device (Table~\ref{tab:devices}).

\textbf{Models and inputs.}
The models cover the forms a looped model takes, from puzzle solvers to an image classifier and a language model. Three are TRM models~\citep{jolicoeur2025trm}. Sudoku-MLP repeats an MLP block (5.0M parameters, counted from the exported graph), and Sudoku-attn and Maze, for $30{\times}30$ mazes, repeat an attention block (6.8M each). HRM is an attention model for Sudoku (27M)~\citep{wang2025hrm}, DeepThinking a convolutional model for $33{\times}33$ mazes (0.78M, counted the same way)~\citep{bansal2022deepthinking}, MDEQ-Small an equilibrium model for ImageNet (18M, as its authors report)~\citep{bai2020mdeq}, and Ouro-1.4B a transformer language model~\citep{zhu2025ouro}. All but DeepThinking and MDEQ-Small have a learned output that says when the answer is ready. Unless stated otherwise, each device answers one input at a time, as a deployed device would. Devices answer the first 100 Sudoku test puzzles and the first 50 test mazes, and the datacenter GPUs answer 1000 puzzles in each timed window. DeepThinking runs 1000 mazes at 10 to 300 iterations. MDEQ runs the first 2500 ImageNet validation images, and Ouro-1.4B runs 30 GSM8K questions in fp16 on the Jetson AGX.

\textbf{Iterations and halting.}
An \emph{iteration} applies the model's repeated block once to its hidden state; for TRM and HRM, it is one high-level cycle. We export each model as a graph of one iteration, so that the hidden state can be read after every iteration. A model with a learned halting output stops an input once that output says its answer is ready, under halting rules fixed before any measurement~\citep{kolawole2026tilted}. An input that does not halt stops at the iteration limit: 48 iterations for the TRM Sudoku models, 32 for HRM and 24 for Maze.

\textbf{Time and energy.}
On 36 devices, Qualcomm AI Hub~\citep{qualcomm2026aihub} compiles each model with QNN, or with TFLite where the device has no NPU path, and reports the kernel time of one iteration. On the other devices, we group three iterations into one call (two for HRM) to spread the launch cost, and divide the time of a call by the iterations it runs. On phones, where we timed both a single iteration and a call of three in 40 cases, the call took a median $2.9$ times as long. Arm's Vela compiler and Fixed Virtual Platform give the time on the Ethos-U NPUs~\citep{arm2026vela,arm2021corstone300}. We measure energy on the Jetson Nano, the Jetson AGX and the L40S, from their power sensors and, on the L40S, NVIDIA's management library, with clocks fixed in the power mode under test. An energy measured this way holds for its power mode and load, so a device whose clocks or load change re-measures one iteration. Each run records an idle window and then five timed windows of 60\,s (120\,s for mazes), and we report the median of the windows that neither throttled nor lost samples. The energy per correct answer (Equation~\ref{eq:jpc}) includes idle power, which a device pays whether or not it computes.

\textbf{Builds and the reference.}
For each device and precision, the toolchain compiles the model into a build, 280 builds in all. We write w4a8 for weights stored in 4 bits and activations in 8; INT8 is w8a8~\citep{nagel2021whitepaper}. Each build runs 8 iterations on the Qualcomm devices, 8 or 16 on the datacenter GPUs (300 for DeepThinking) and 16 on the Jetsons and the Raspberry Pi, and keeps the answer and the hidden state after each. The fp32 reference of Section~\ref{sec:background} runs on a host computer with ONNX Runtime, or as the board's own fp32 build on the Jetsons, the Raspberry Pi and the server CPU.

\textbf{Presentation of results.}
Accuracy is the fraction of inputs answered correctly. A prediction's error is its relative difference from the measured value, and we report the median over a device's runs. Accuracy curves over iterations come from the host's fp32 run unless a figure names a device.

\section{The energy and count of iterations}
\label{sec:characterize}

A developer deploying a looped model needs to know what an answer will cost on a device and how accurate it can be there. In this section, we show that both follow from measuring a single iteration on the device. We first price every run from its energy (\S\ref{sec:cost}), then read the accuracy within a latency budget from its time (\S\ref{sec:predict}), and finally show where halting lowers the count (\S\ref{sec:halting}). Section~\ref{sec:correct} turns to correctness, the third property of iteration.

\subsection{A constant energy per iteration}
\label{sec:cost}

\textbf{The energy of a single iteration is enough to price every run on a device.}
The hidden state has the same size at every iteration, so an iteration costs the same energy $e_d$ each time it runs on device $d$, at a given precision and power mode. A run with a mean of $n$ iterations per input and accuracy $\mathrm{acc}(n)$ then costs
\begin{equation}
E_d(n) = \frac{n\, \textcolor{amberdark}{e_d}}{\textcolor{hostteal}{\mathrm{acc}(n)}}
\label{eq:cost}
\end{equation}
per correct answer, with no fixed cost per input, such as loading it. In words, the energy a run spends on its iterations is spread over the inputs it answers correctly: more iterations raise the cost in proportion, and more correct answers lower it. Equivalently, $E_d(n) = \textcolor{amberdark}{e_d}/\textcolor{hostteal}{\mathrm{IPI}(n)}$: the device sets the energy of an iteration, and the model sets its intelligence per iteration (\S\ref{sec:background}).

\textbf{The median prediction error is $0.5\%$ on the Jetson AGX and at most $8\%$ on any device.}
To test the rule, we take $e_d$ from one run of at most 16 iterations for each model and configuration, and predict every other run's energy per correct answer from its mean iteration count and accuracy. Figure~\ref{fig:energy} shows that the predictions fall on the measured values, for runs of 3 to 96 iterations. The predictions hold on the Jetson AGX and the L40S alike, on the L40S even for runs six times as long as the one that set $e_d$ (median error $1.7\%$). The Jetson Nano's error is higher because its timed windows score different subsets of the puzzles, so its accuracy varies between windows while the energy of an iteration barely does. The rule also holds for runs that end with higher-precision iterations, such as the repaired builds of Section~\ref{sec:repair}: on the Jetson AGX, two such runs fall within $5.3\%$.

The rule extends to a looped language model. On the Jetson AGX, Ouro-1.4B in fp16 spends the same energy per token on each loop it runs, from one loop to four, as Equation~\ref{eq:cost} assumes of an iteration. Priced from its 1-loop run alone, its energy at 2 to 4 loops falls within $4.1\%$ of the measured value. As more loops raise its accuracy on GSM8K fourfold, its energy per correct answer falls by about a quarter, because its answers also take a fifth fewer tokens.

\begin{figure}[t]
\centering
\includegraphics{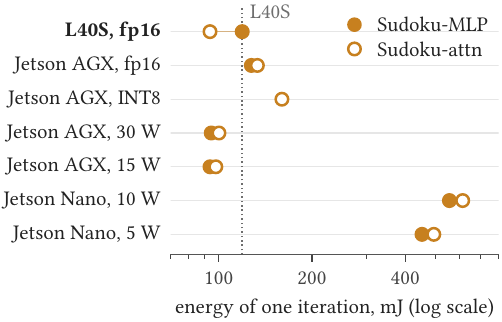}
\caption{\textbf{At batch size 1, a Sudoku-MLP iteration costs 127\,mJ on a Jetson AGX and 119\,mJ on an L40S.} TensorRT throughout. In fp16, a Sudoku-attn iteration costs 1.4$\times$ as much on the Jetson AGX as on the L40S, and one of the larger Maze model, not shown, 3.7$\times$.}
\label{fig:itercost}
\end{figure}

\textbf{FLOP counts miss the energy of an iteration by up to $97\times$.}
To estimate the energy of an iteration from FLOPs, we divide its FLOPs by the device's peak throughput and multiply by its rated power. Figure~\ref{fig:energy} shows that this estimate falls short of the measured energy for every model and runtime. No constant factor corrects the estimate, because the ratio of measured to estimated energy changes by more than an order of magnitude across runtimes on the same L40S. The energy depends on how the runtime executes an iteration, which a FLOP count cannot see. Prior work reports such shortfalls for networks that run once~\citep{yang2017energyaware,fernandez2025energy}; we measure them per iteration, across runtimes and devices.

\textbf{At batch size 1, an L40S spends as much per iteration as a Jetson AGX.}
Measuring the iteration also overturns a common expectation about where a model should run. A datacenter GPU is efficient when it serves many inputs at once, whereas a device answers one at a time. Yet at batch size 1, an iteration of Sudoku-MLP costs 119\,mJ on the L40S and 127\,mJ on the Jetson AGX in fp16 (Figure~\ref{fig:itercost}). The L40S draws about $7\times$ the power of the Jetson AGX for about an eighth of the time, so it runs the model far faster at about the same energy. Intelligence per watt finds local accelerators at least $1.4\times$ behind cloud ones at batch size 1~\citep{saadfalcon2025ipw}; in energy per iteration, by contrast, the Jetson AGX comes within $7\%$ of the L40S.

Lower precision does not reliably cut the energy of an iteration. One would expect an INT8 build to cost less than an fp16 one. At batch size 1, however, TensorRT on the Jetson AGX finds every matrix multiply faster in fp16 than in INT8, so its INT8 builds run them in fp16. Its INT8 builds therefore cost $3$--$14\%$ more energy per solved puzzle than its fp16 builds. On phones, the w8a16 Sudoku builds of the Galaxy S24 and S25 run as fast as their w8a8 ones, while the S25's w8a8 Maze build runs $2.4\times$ faster than its fp16 build. Whether lower precision pays off thus depends on the model and the device, and we learn which by timing a single iteration of each build.

\textbf{The cheapest iteration count is the same on every device.}
Since the energy of an iteration differs by up to $4.4\times$ across the boards and power modes on which we ran several fixed counts (Figure~\ref{fig:itercost}), each device might seem to need its own iteration count. The count at which an answer is cheapest, however, maximizes the intelligence per iteration (Figure~\ref{fig:cost}). Because $e_d$ only scales Equation~\ref{eq:cost}, this maximum depends on the accuracy curve alone, which a correct build shares with the host. Among the fixed counts we ran on each board (4, 8 and 16), the cheapest was the same, 4 iterations, for both Sudoku models on the Jetson Nano in both power modes and on the Jetson AGX. We therefore set the count once, on the host, for every device; what differs between devices is how many iterations fit within a latency budget (\S\ref{sec:predict}).

\subsection{Accuracy within a latency budget}
\label{sec:predict}

\begin{figure}[t]
\centering
\includegraphics{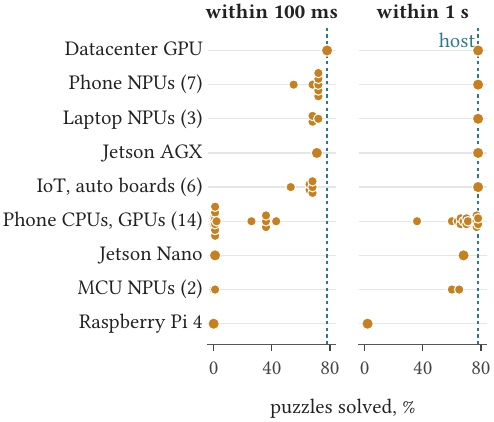}
\caption{\textbf{Within 100\,ms, the same Sudoku weights solve 53--78\% of puzzles on 18 devices and at most 2\% on 13 others.} One dot per device: the host's Sudoku-MLP accuracy after as many iterations as fit in the budget at that device's measured time per iteration.}
\label{fig:ladder}
\end{figure}

\textbf{The host's curve and the time of a single iteration are enough to give a device's accuracy within any budget.}
Within a budget, a device runs as many iterations as fit, a number set by its time per iteration (Table~\ref{tab:devices}). A correct build then answers as accurately as the host does after that many iterations, so the host's accuracy curve gives the device's. We check this on the builds for Qualcomm devices, whose accuracy we recorded after every iteration. Quantized builds that are correct against the host fp32 run differ from its accuracy by 3.1 percentage points on average over iterations. The accuracy predicted within a budget is therefore within the same 3.1 points.

\textbf{Within 1\,s, the same Sudoku weights solve $2\%$ to $78\%$ of puzzles.}
Figure~\ref{fig:ladder} shows where each device lands, set by its time per iteration. Within 1\,s, the fastest devices, among them the Jetson AGX and two phones, run enough iterations to reach the host's final accuracy, while the Raspberry Pi~4 runs only two. DeepThinking and MDEQ spread as widely. Within 250\,ms, DeepThinking solves every maze on three Snapdragon devices and almost none on more than half of the rest. Within 100\,ms, MDEQ classifies $78\%$ of images on half the devices and under a fifth on the others. A developer can thus choose a device for a budget from the time of a single iteration, without running every iteration count there.

\subsection{Savings from halting}
\label{sec:halting}

\begin{figure}[t]
\centering
\includegraphics{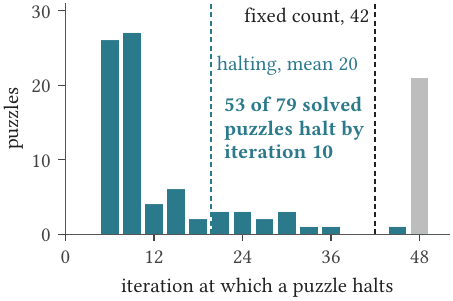}
\caption{\textbf{Most puzzles halt long before the fixed count, so halting cuts the energy per correct answer by $2.1\times$ at equal accuracy.} Sudoku-MLP on the Jetson AGX: when each of 100 puzzles halts, in grey for puzzles it does not solve. The fixed count is the cheapest one that matches halting's accuracy.}
\label{fig:halting}
\end{figure}

\textbf{Halting cuts the energy per correct answer by up to $2.3\times$.}
So far, every input in a run received the same number of iterations. Intuitively, halting saves energy because the cost of a run grows with its mean iteration count (Equation~\ref{eq:cost}), so a run that reaches an accuracy in fewer iterations per input costs less. We compare halting with the best fixed count of Section~\ref{sec:background}, chosen with the test labels. No developer can set a better count without labels, so these savings are the smallest that halting offers. HRM's authors compared halting with a fixed count at its maximum~\citep{wang2025hrm}; we compare with the best fixed count and measure the energy on devices. On the Jetson AGX, Sudoku-MLP halts after about half as many iterations, on average, as the fixed count needs, and its saving, $2.1\times$, equals the ratio of the two counts, as Equation~\ref{eq:cost} predicts at equal accuracy. Across devices, halting saves about $2\times$ for both Sudoku models.

\textbf{The saving comes from inputs that halt early.}
A fixed count runs every input as long as the slowest input it must solve. On the Jetson AGX, two thirds of the puzzles that Sudoku-MLP solves halt within a quarter of the fixed count's iterations (Figure~\ref{fig:halting}). Meanwhile, inputs that halting does not solve run to the limit and cost energy without a correct answer, as every unsolved Sudoku-MLP puzzle does.

The saving shrinks where a fixed count reaches halting's accuracy early. HRM halts after about two thirds of its iteration limit on every device. On the Jetson Nano, a fixed count reaches the same accuracy as early, so halting saves nothing. On the L40S's 1000 puzzles, by contrast, the fixed-count accuracy keeps rising to the limit, and halting saves $1.6\times$. Likewise, every maze that the Maze model solves halts at 6 iterations, the earliest it can, so a fixed count of 6 matches it on the Jetsons. Among the L40S's 1000 mazes, however, some solved mazes need more iterations, so a fixed count needs twice as many and halting saves $2.0\times$. How early inputs halt and how many reach the limit are properties of the model and its inputs, which we measure on the host before deployment.

\begin{takeaway}{amber}{amberfill}{One iteration prices every run}
Every iteration repeats the same computation, so on a given device each one costs the same energy and time. The energy of a single iteration thus predicts the cost of every run to a median error of at most 8\%, where FLOP estimates fall short by up to 97$\times$, and its duration gives the accuracy within any latency budget.
\end{takeaway}

\section{Silent faults, and a check in two iterations}
\label{sec:correct}

Section~\ref{sec:characterize} assumed that the device computes each iteration correctly. In this section, we show that it often does not, and that the device can tell. We first show that builds from every toolchain we checked run without error and compute wrong answers (\S\ref{sec:faults}), then explain why their error shows within two iterations (\S\ref{sec:early}). The device can therefore check its build without labels (\S\ref{sec:check}), and repair a build that fails by running a few more iterations at higher precision (\S\ref{sec:repair}).

\subsection{Builds that fail without an error}
\label{sec:faults}

\textbf{94 of the 280 builds run without error and compute wrong answers.}
All 280 builds compiled and ran to completion without an error, under QNN, TensorRT and ONNX Runtime alike. Some fail completely: a w4a16 Maze build on a Snapdragon NPU solves no mazes, where the host solves $88\%$, and an INT8 DeepThinking build under TensorRT solves none either, where the host solves every maze. Their outputs give no sign of the fault. The broken phone builds solve no input, yet get 32--90\% of individual cells right, so their output looks like a hard input answered wrongly. Only TFLite Micro fails visibly. Each of the 14 quantized builds it runs for the Ethos-U returns one constant output for every input, from a single fault in its reference kernels; we count these apart from the 280.

\textbf{The toolchains' standard builds break as well.}
A developer who takes a toolchain's standard path meets these faults. The standard w8a8 Maze build is broken on 10 of 13 Snapdragon devices. On the datacenter GPUs, every explicit-INT8 build, whose graph carries its own quantization steps, is broken (24 of 24), while TensorRT's implicit INT8 builds of the same models are correct. For DeepThinking and MDEQ, ONNX Runtime fails on the same explicit graph, indicating that the fault arises from the graph's quantization steps, which every runtime then executes. On the Jetson AGX, constraining every matrix multiply and convolution to INT8 breaks the Maze build as well.

\textbf{Simulating a build on the host passes most broken phone builds.}
Of the 72 phone builds we hold out to test the check (\S\ref{sec:check}), 47 are broken. For 30 of the 46 we simulated, the quantized graph is correct on the host, and for 5 more it falls only a few puzzles short, so the device's toolchain introduced the error. A developer who validated these builds by simulating them would ship them. The other 11 fail in simulation too, yet one of their graphs runs correctly on nearly every other device it is compiled for, so whether a graph works depends on the device.

\subsection{Why the error shows within two iterations}
\label{sec:early}

\begin{figure}[t]
\centering
\includegraphics{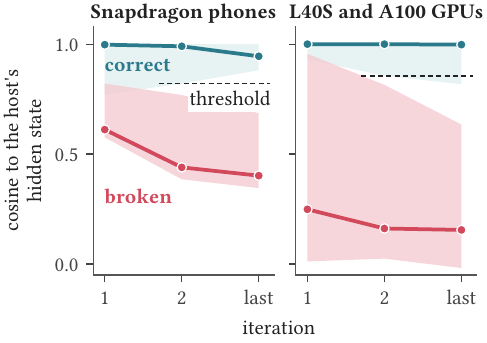}
\caption{\textbf{After two iterations, every broken build's hidden state is farther from the host's than any correct build's.} Cosine with the host's fp32 hidden state, degraded builds left out. Bands span all builds and lines are medians; dashed lines are each toolchain's threshold (\S\ref{sec:check}).}
\label{fig:check}
\end{figure}

\textbf{A broken build leaves the host's path within two iterations and does not return.}
To see when a broken build goes wrong, we compare its hidden state after every iteration with that of the host's fp32 run, by their cosine similarity. On the datacenter GPUs, after the second iteration, the hidden state of a typical broken build is nearly orthogonal to the host's, while that of a correct build nearly matches it (median cosine $0.15$ against $1.00$). The two groups stay apart until the last iteration. Figure~\ref{fig:check} shows the same split on the phone builds that set the check's threshold (the threshold builds, \S\ref{sec:check}) and on every GPU build: after the second iteration, every correct build stays above every broken one.

Intuitively, a correct quantized build applies a slightly perturbed copy of the host's iteration, so each iteration adds a small error and its state stays near the host's path. By contrast, a broken build applies a different function at every iteration, each time to a state already off the host's path, so its state never returns to that path. After one iteration, we can already separate most broken phone builds from correct ones by their hidden state, at an area under the receiver operating characteristic curve (AUROC) of 0.972; after two, we can separate all of them.

\textbf{Iteration carries a fault forward.}
A looped model runs the same operations at every iteration and passes its hidden state to the next, so a faulty operation errs at every iteration and each error carries forward. The faults we diagnosed sit in specific operations: a w8a8 Maze build fails on two NPU generations until its rotary position embedding runs in fp16, and the w8a8 Sudoku-MLP graph solves no puzzles with its carried state in 8 bits but $55$--$61\%$ with it in fp16.

This mechanism explains the rest of the section. A broken build's output looks plausible because its iterations still move the state toward an answer, so many cells come out right (\S\ref{sec:faults}). Its outputs therefore separate builds less well than its hidden state does (\S\ref{sec:check}). The reference must be the fp32 run, because a host simulation of the quantized graph shares the graph's own faults and misses the device's. A broken build can also be repaired, because a few correct iterations can finish the path that its own iterations began (\S\ref{sec:repair}).

\subsection{A check in two iterations}
\label{sec:check}

\begin{table}[t]
\centering
\caption{\textbf{On builds it never saw, the hidden state after two iterations misses none of 63 broken builds, while output agreement misses 47.} Broken builds missed at a threshold set on other builds; no signal flags a correct build. \nd: not recorded for that set.}
\label{tab:check}
\setlength{\tabcolsep}{3pt}
\begin{tabular}{@{}>{\raggedright\arraybackslash}p{4.35cm}rr@{}}
\toprule
Signal & phones, of 47 & A100, of 16 \\
\midrule
\rowcolor{hostfill}\textbf{Hidden state, iteration 2} & \textbf{0} & \textbf{0} \\
Hidden state, iteration 1 & \nd & \textcolor{badred}{\textbf{2}} \\
Output agreement, iteration 2 & \textcolor{badred}{\textbf{39}} & \textcolor{badred}{\textbf{8}} \\
Output agreement, iteration 1 & \textcolor{badred}{\textbf{47}} & \nd \\
Projected distance, iteration 8 & 0 & \nd \\
\bottomrule
\end{tabular}
\end{table}

\textbf{The hidden state after two iterations is enough to separate every broken build from every correct one.}
On the device, the check runs the build for two iterations and compares its hidden state with that of the host's fp32 run, by their cosine similarity averaged over inputs (Figure~\ref{fig:check}). A build fails when this cosine falls below a threshold. We set the threshold once per toolchain, at the lowest cosine of any correct build in a set whose accuracy we know. Labels are needed only on the host, for these builds: the threshold builds set the phone threshold at 0.821, and the L40S's builds set the GPU threshold at 0.856.

\textbf{The thresholds carry to new models, precisions and devices with no miss and no false alarm.}
Applied unchanged to 72 further phone builds, the phone threshold catches all 47 broken builds and flags none of the 24 correct ones (Table~\ref{tab:check}). Most of these builds come from graphs the threshold never saw, among them a fourth model and three new precisions. A third of them run on devices that no threshold build ran on. On an A100, the L40S's threshold catches all 16 broken builds and flags none of the 28 correct ones. Degraded builds compute what correct builds of the same graph compute: their hidden states match to the third decimal, and their accuracy falls a few puzzles short of the $95\%$ line. The check therefore groups them with correct builds.

Simpler signals fall short of the hidden state after two iterations. After one iteration, the hidden state still misses 2 of the 16 broken A100 builds (Table~\ref{tab:check}). At a threshold carried over in the same way, agreement of the outputs with the host's after two iterations misses 39 of the 47 broken held-out phone builds and 8 of the 16 broken A100 builds, because broken builds get most cells right (\S\ref{sec:faults}, Table~\ref{tab:check}). A constant-output test, which catches the microcontrollers' fault, scores 0.500, because broken phone builds vary their output with the input.

The check owes its accuracy to its fp32 reference and its second iteration more than to the distance it uses. DiFR's distance between random projections of the two hidden states~\citep{karvonen2025difr} also separates the held-out phones after two iterations, and after eight it catches every broken one at a threshold from the threshold builds (Table~\ref{tab:check}). Against a host simulation of the same quantized graph, however, the hidden state scores an AUROC of only 0.756 on the threshold builds. For the w8a8 Sudoku-MLP graph of Section~\ref{sec:early}, the simulated graph was itself the broken one. Measured against such a reference, a working device would fail the check, and a fault in the graph itself would pass.

\subsection{Repair on the device}
\label{sec:repair}

\begin{figure}[t]
\centering
\includegraphics{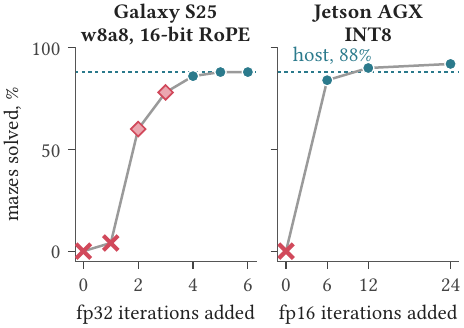}
\caption{\textbf{A few higher-precision iterations appended to a broken build's run take it from 0\% to the host's accuracy or above.} Maze, 50 test mazes; each point is coloured by the build's verdict at that step. The AGX build forces every matrix multiply and convolution to INT8.}
\label{fig:repair}
\end{figure}

\begin{figure*}[t]
\centering
\input{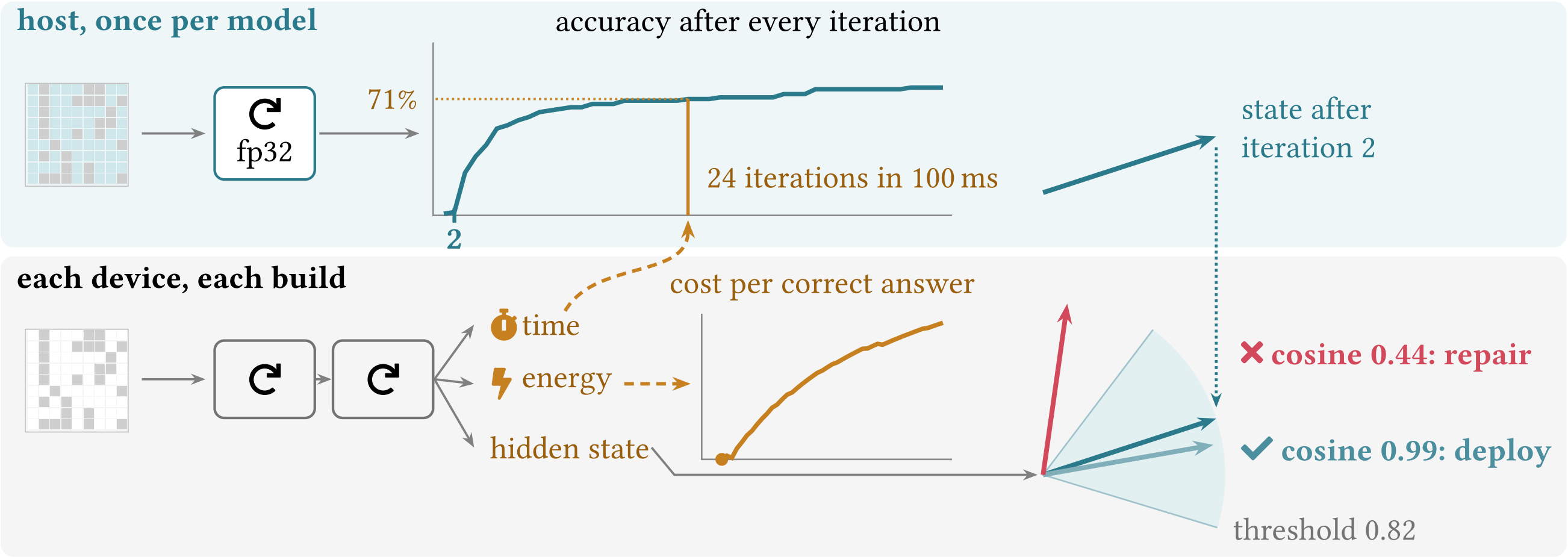}
\caption{\textbf{\methodname{} runs the model once on the host and two iterations on each device, and reads the rest off the host's run.} Sudoku-MLP on a Jetson AGX: an iteration's time sets where the host's curve is read and its energy prices every answer; the cosines drawn are the medians of correct and broken phone builds.}
\label{fig:procedure}
\end{figure*}

\textbf{Running a few more iterations at higher precision restores a broken build.}
As Section~\ref{sec:early} anticipates, a build that fails the check can be repaired: it runs as usual and then runs a few more iterations at higher precision on the same device (Figure~\ref{fig:repair}). On the Galaxy S25, a w8a8 Maze build that keeps its rotary position embedding at 16 bits solves no mazes, although it gets $90\%$ of their cells right. With two fp32 iterations appended to its run, it solves five times as many mazes as a fresh two-iteration fp32 run, and with five, it reaches the host's $88\%$. Repair needs only a higher-precision build of the same graph on the device, and no labels.

\textbf{Repair can surpass the higher-precision build itself.}
On the Jetson AGX, repair takes an INT8 Maze build that solves no mazes to $92\%$, above the $88\%$ that the fp16 build reaches at any iteration count we ran, and to $84\%$ after only 6 fp16 iterations. Repair also raises a degraded build: an INT8 Sudoku-attn build rises from $55\%$ to $70\%$.

The repair itself costs little. Per maze, six fp16 iterations add 5.9\,J and 159\,ms on the Jetson AGX, and five fp32 iterations add 224\,ms on the S25. A repaired run still costs more than the higher-precision build alone, because of the iterations before repair. First, a broken build rarely halts, so it runs to its limit before repair begins. On the S25, these iterations, though faster than fp32 ones, together carry the state only about as far as a single fp32 iteration would, so the repaired run takes about $1.5\times$ as long as an fp32 run. Second, on the Jetson AGX, the INT8 build is slower per iteration than the fp16 one, so running it first saves nothing, and the repaired run spends seven to nine times the fp16 build's energy per solved maze. Repair therefore pays off in energy where the low-precision build is faster and halts on its own. Elsewhere, a failed check points to deploying the higher-precision build instead.

\begin{takeaway}{hostteal}{hostfill!70}{Two iterations check a build}
A faulty build repeats its error at every iteration, so its hidden state leaves the host's within two iterations, while a correct build's stays close. On held-out builds, we catch all 63 broken builds this way and flag no correct one, without labels. Running a few more iterations at higher precision then restores a broken build.
\end{takeaway}

\section{\methodname{}: measure once, check every device}
\label{sec:procedure}
\label{sec:procedure-steps}

To deploy a looped model across devices, \methodname{} turns the findings of Sections~\ref{sec:characterize} and~\ref{sec:correct} into a procedure. It has two parts: a host part, run once per model, and a device part, run on each build before it is deployed (Figure~\ref{fig:procedure}). We follow the Maze model through both.

\textbf{The host runs the model once, for every device.}
In fp32, on a set of inputs, the host records the accuracy after every iteration, the iteration at which each input halts, and the hidden state after iteration 2. These depend only on the model and its inputs, not on the device. The host also sets the check's threshold once per toolchain, on builds whose accuracy it knows, since builds from different toolchains depart from fp32 by different amounts (\S\ref{sec:check}). For the Maze model, the host's run solves $88\%$ of mazes.

\textbf{Each device runs its build for two iterations, on the same inputs and without labels.}
We compare the hidden state after the second iteration with the host's, since by then every broken build has left the host's path (\S\ref{sec:early}). On the Galaxy S25, a w8a8 Maze build with its rotary embedding at 16 bits runs these iterations at 17.8\,ms each, and its hidden state then has a cosine of 0.77 with the host's, below the phone threshold of 0.821. The build fails, and is repaired by running a few more iterations at higher precision after its run, since its own iterations still move the state toward the answer (\S\ref{sec:repair}). With five fp32 iterations at the end of its run, at 45\,ms each, it reaches the host's $88\%$, and the check and repair together take about 260\,ms per input.

On the Jetson AGX, by contrast, the fp16 Maze build passes. We time an iteration in the same run and read the accuracy within any latency budget from the host's curve (\S\ref{sec:predict}). At 26\,ms per iteration, all 24 iterations fit within 1\,s, so the build reaches the host's $88\%$ in that budget. On a device with power sensors such as this one, a short timed run gives the energy of an iteration, from which we price every answer (\S\ref{sec:cost}). Because a lower precision is not always faster (\S\ref{sec:cost}), the device also times one iteration of the build at the other precision, and keeps whichever correct build costs less. On the Jetson AGX, this is the fp16 build, since INT8 runs slower there.

\textbf{The check costs each device two iterations per input.}
On the slowest device, the Raspberry Pi~4, two iterations take under 1\,s per input at 378\,ms each; on the Snapdragon phones' NPUs, they take 6 to 25\,ms per Sudoku input (Table~\ref{tab:devices}). Pricing adds one short timed run on devices with power sensors, and one timed iteration of the higher-precision build. Without the procedure, a device would need a labelled run to the iteration limit for every build to learn its accuracy curve. Two iterations are $24\times$ fewer than such a run for the 48-iteration Sudoku models, and $150\times$ fewer for DeepThinking's 300.

\textbf{Toolchains can run the check at compile time.}
A toolchain reports only that a build runs (\S\ref{sec:intro}), yet the check needs only the hidden states of one fp32 run on a few inputs. A toolchain, or a service such as Qualcomm AI Hub that already runs inference on its devices, can therefore compare a build's hidden states with the fp32 run's when it compiles the build, and refuse a broken build before it ships.

\section{Discussion and conclusion}
\label{sec:discussion}

Looped models give edge devices more accuracy without more memory, but a developer must know what each device will deliver before deploying one. In this work, we showed that a few iterations on each device are enough to find out. Since every iteration repeats the same computation, the energy of one iteration prices every run on a device and its time gives the accuracy within any latency budget, which FLOP estimates cannot do. Each iteration also carries its state forward, so a broken build departs from the host's run within two iterations, and we catch it there without labels. \methodname{} combines these findings to price and check the build on each device, and repairs a broken one with a few higher-precision iterations. We list the main implications of this work and the questions it leaves open below.

\textbf{Benchmarks should report a curve and a check.}
Current benchmarks such as intelligence per watt score each model and device at one setting, from the device's own outputs~\citep{saadfalcon2025ipw}. For a looped model, that setting is one point on a curve over iterations, and the latency budget and the halting rule decide which point a device reaches (\S\ref{sec:predict}). Scoring the device's outputs also charges a broken build's errors to the model or the device (\S\ref{sec:faults}). A benchmark of looped models should therefore report the energy per correct answer at every iteration count, and check each build against a reference run.

\textbf{The findings hold for any model that repeats a fixed-size step.}
Since Equation~\ref{eq:cost} and the check need only a step that repeats on a state of fixed size, they apply to the looped and equilibrium models we measured, and the equation also applies to diffusion models. Where the state grows, as in the search and agent loops that language models run on devices~\citep{hao2025ttsnpu,tian2026agentsedge}, the equation needs a cost per step in place of $e_d$. Where each input carries a fixed cost, such as encoding a prompt, the equation needs a term for it as well. For the check, whether two steps reveal broken diffusion builds remains open.

\textbf{Looped language models are the next test.}
Several open-weight language models now run their layers more than once per token~\citep{nanbeige2026loop,sensenova2026loop,tml2026inkling2mini}. However, prior work serves them~\citep{du2026loopduet} or shrinks their memory~\citep{vendrell2026melt} on datacenter GPUs, without measuring their energy. On the Jetson AGX, we find that each of Ouro's loops costs the same energy, so Equation~\ref{eq:cost} prices the loops of a token (\S\ref{sec:cost}). Across tokens, by contrast, the cache grows, so pricing a whole answer needs a cost per step, unless eviction holds the cache to a fixed budget~\citep{gao2025kvrethink}. Extending the price of an iteration to whole answers, and the check to looped language models' builds, would carry these findings to the models now shipping for edge devices.

\bibliographystyle{ACM-Reference-Format}
\bibliography{references,refs_lead,refs_w1,refs_w2,refs_w3}
\end{document}